\documentclass[10pt,twocolumn,letterpaper]{article}

\usepackage[pagenumbers]{cvpr} 

\usepackage{graphicx}
\usepackage{amsmath}
\usepackage{amssymb}
\usepackage{booktabs}
\usepackage{caption}
\usepackage{subcaption}
\usepackage[pagebackref,breaklinks,colorlinks]{hyperref}

\usepackage[capitalize]{cleveref}
\crefname{section}{Sec.}{Secs.}
\Crefname{section}{Section}{Sections}
\Crefname{table}{Table}{Tables}
\crefname{table}{Tab.}{Tabs.}

\begin{document}

\title{OmniDataComposer: A Unified Data Structure for Multimodal Data Fusion and Infinite Data Generation}

\author{Dongyang Yu Email: \href{mailto:yudongyang2022@gmail.com}{yudongyang2022@gmail.com}Shihao Wang Email: \href{wanghao.cst@gmail.com}{wanghao.cst@gmail.com}\\Yuan Fang Email: \href{mailto:ryanfang.cs@gmail.com}{ryanfang.cs@gmail.com}Wangpeng An Email: \href{mailto:anwangpeng@gmail.com}{anwangpeng@gmail.com}}

\maketitle

\begin{abstract}

This paper presents OmniDataComposer, an innovative approach for multimodal data fusion and unlimited data generation with an intent to refine and uncomplicate interplay among diverse data modalities. Coming to the core breakthrough, it introduces a cohesive data structure proficient in processing and merging multimodal data inputs, which include video, audio, and text.

Our crafted algorithm leverages advancements across multiple operations such as video/image caption extraction, dense caption extraction, Automatic Speech Recognition (ASR), Optical Character Recognition (OCR), Recognize Anything Model(RAM), and object tracking. OmniDataComposer is capable of identifying over 6400 categories of objects, substantially broadening the spectrum of visual information. It amalgamates these diverse modalities, promoting reciprocal enhancement among modalities and facilitating cross-modal data correction. \textbf{The final output metamorphoses each video input into an elaborate sequential document}, virtually transmuting videos into thorough narratives, making them easier to be processed by large language models.

Future prospects include optimizing datasets for each modality to encourage unlimited data generation. This robust base will offer priceless insights to models like ChatGPT, enabling them to create higher quality datasets for video captioning and easing question-answering tasks based on video content. OmniDataComposer inaugurates a new stage in multimodal learning, imparting enormous potential for augmenting AI's understanding and generation of complex, real-world data.

\end{abstract}

\begin{figure*}[ht]
  \centering
  \includegraphics[height=6in]{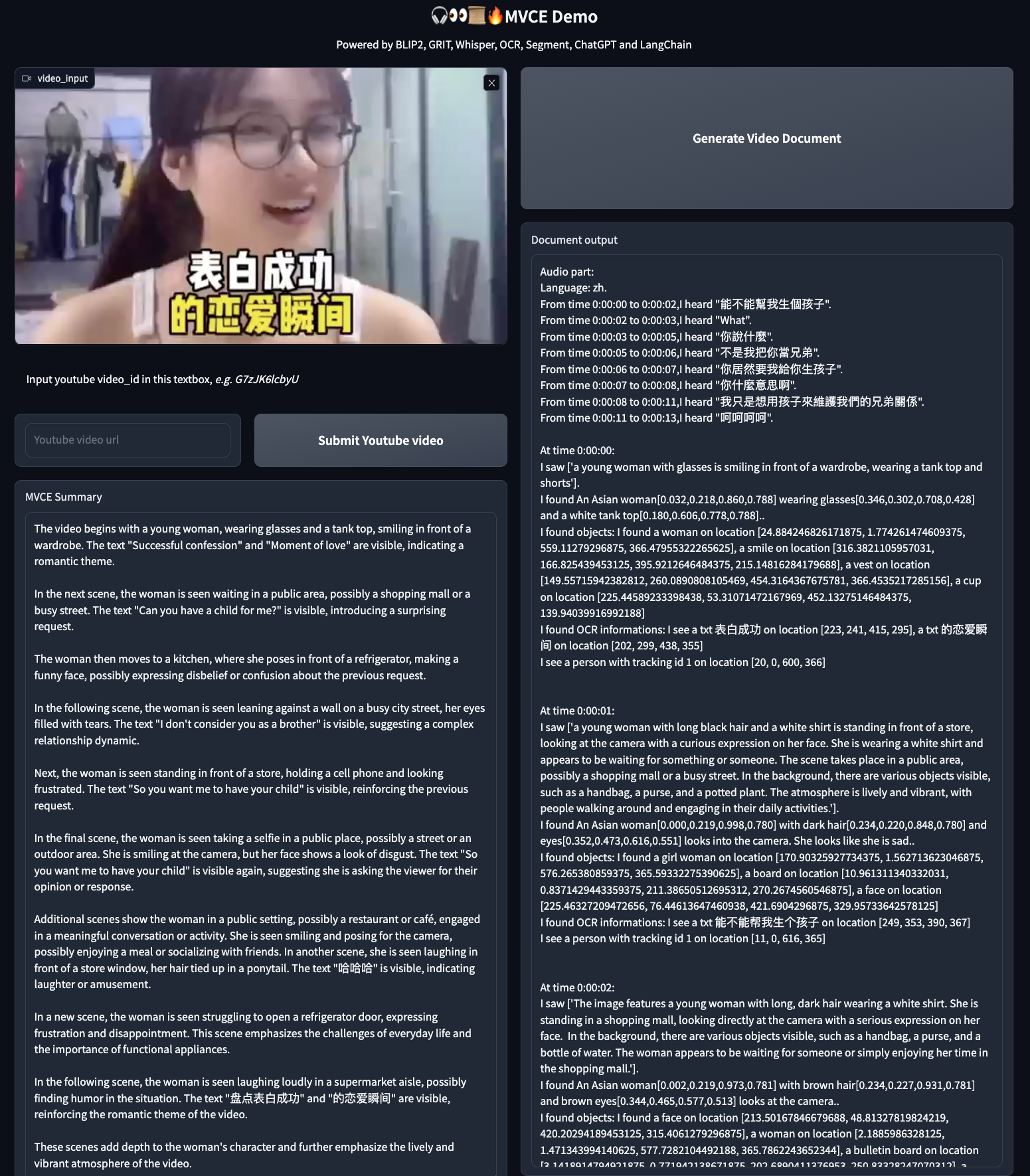}  

  \caption{Demo on Input Video}
  \label{fig:overview1}
\end{figure*}

\section{Introduction}
In the rapidly evolving landscape of artificial intelligence and machine learning, the ability to efficiently process and make sense of diverse data types is paramount. The increasing availability of vast amounts of data in different formats, including text, images, audio, and video, necessitates the development of robust models capable of handling such data in an integrated manner. This has given rise to the field of multimodal learning, where the goal is to build models that can understand, learn from, and generate outputs across multiple data modalities. 

Despite significant progress in individual modalities, current solutions often operate in isolation, with little cross-modal interaction and correction. This leads to a disjointed understanding and use of data, restricting the full exploitation of information-rich environments. Furthermore, while considerable progress has been made in converting image into text \cite{alayrac2022flamingo,li2022blip,dai2023instructblip,gong2023multimodal,wang2023visionllm,liu2023visual}, audio into text \cite{radford2023robust,bain2023whisperx,gong2023whisper} and video into text \cite{videochatlong,li2023videochat,wang2023chatvideo}, these are often processed separately, without a cohesive framework to combine these modalities and extract maximum information.

Our paper presents OmniDataComposer, an innovative approach to unifying diverse data modalities into a single, coherent data structure. This algorithm extracts valuable information from video through captioning, ASR, OCR, and object tracking, subsequently integrating these different modalities in a mutually supportive and corrective manner. The outcome is a detailed sequential long document for each video input, providing a rich, comprehensive narrative of the video content.
Fig.~\ref{fig:overview1} provides an illustration of a demo that we have developed to process videos, showcasing the operational workflow employed by OmniDataComposer.
Beyond this, we empower OmniDataComposer to serve as a robust foundation for infinite data generation and optimization of datasets for each modality. This lays the groundwork for a new era of multimodal learning, providing large language models like ChatGPT with the capacity to generate high-quality video captioning datasets and facilitating question-answering tasks based on video content.

This paper is structured as follows: we first discuss related works in multimodal learning, video processing, and large language models. Next, we detail the OmniDataComposer algorithm, followed by an exploration of our experimental results and ablation studies. Finally, we conclude with future directions for this work.

\section{Related Works}
The field of multimodal learning has garnered significant attention in recent years due to the immense value it brings to understanding and processing diverse and complex data types. However, individual areas of research that contribute to this overarching theme, such as video processing, audio processing, and large language models, each possess their own nuances and challenges.
\subsection{Video Processing}
The transformation of video data into a more accessible format like text has been a key area of interest. Work on video captioning, like the S2VT model \cite{venugopalan2016improving} and others \cite{li2022blip,li2023blip2}, propose methods to generate descriptions for short video clips. Meanwhile, research on object detection and tracking within video data has made considerable strides, with models such as YOLO and DeepSORT \cite{wojke2017simple} leading the field. However, these works tend to process information from each frame independently, missing temporal dependencies.

\begin{figure*}[ht]
\centering
\includegraphics[width=6.5in]{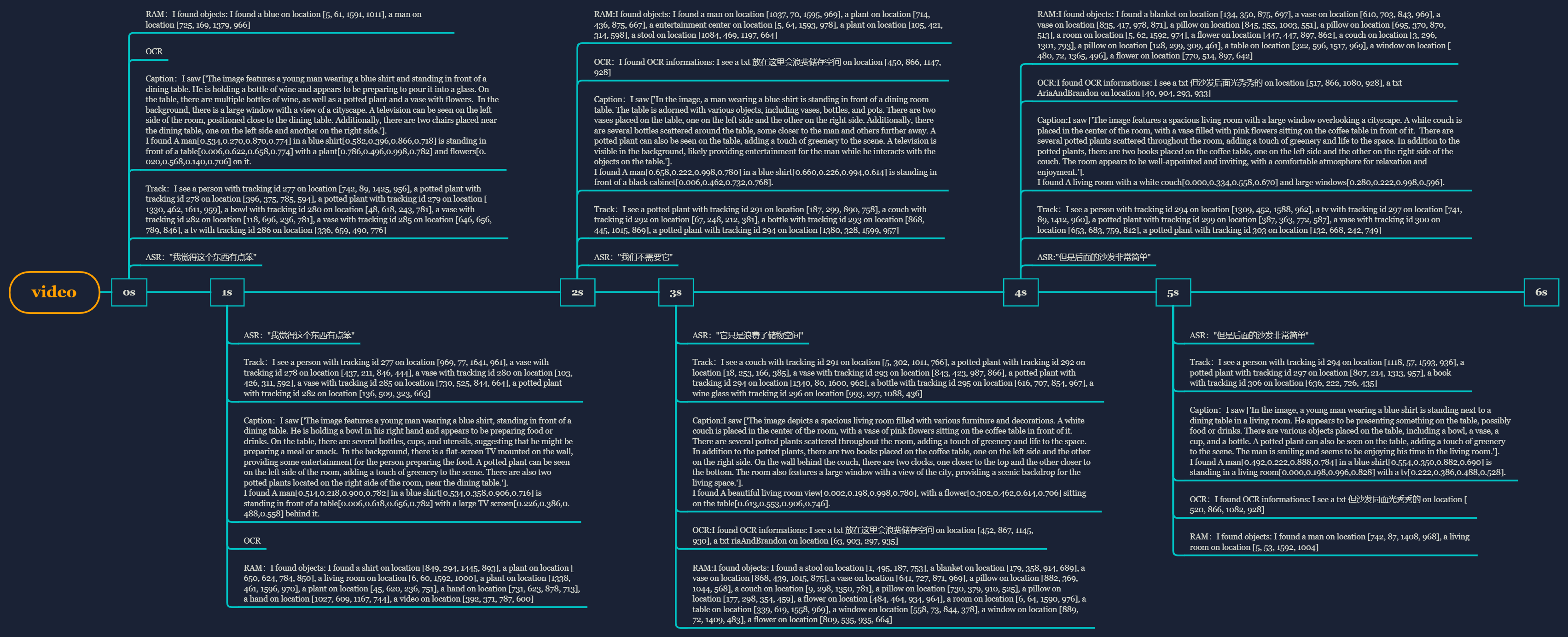}
\caption{Overview of OmniDataComposer Algorithm}
\label{fig:Algorithm}
\end{figure*}

\subsection{Audio Processing}

In parallel, automatic speech recognition (ASR) has been instrumental in converting spoken language into text. Whisper \cite{gong2023whisper} is a notable instance of such technology. However, ASR models have typically been used in isolation without strong integration with other modalities.
\subsection{Optical Character Recognition}
Significant progress has been made in the field of Optical Character Recognition (OCR), which is a key element in extracting textual information from images or videos. Models like MaskOCR \cite{lyu2022maskocr} are excellent at text detection and recognition. Yet, like other areas, this modality is often processed separately from others.
\subsection{Large Language Models}
The advent of transformer-based models like BERT \cite{kenton2019bert}, GPT4 \cite{gpt4}, and CLIP \cite{radford2021learning} has revolutionized language understanding and generation tasks. While CLIP demonstrated powerful zero-shot capabilities by jointly learning from images and text, the joint processing of video, audio, and text is still largely unexplored.

LLaVa \cite{liu2023visual} showcases the effectiveness of employing language-only GPT-4 for visual instruction tuning and introduces a data pipeline to generate image-level instruction-following data. However, when it comes to achieving video-level comprehension, its performance is constrained by the quantity of modalities available.

Our work, OmniDataComposer, leverages the strengths of the individual advancements in these fields and brings them together to form a cohesive and comprehensive approach. By creating a unified data structure that fuses video, audio, and textual information, we aim to overcome the limitations of disjointed processing, bringing forth richer, more contextually aware outputs.

\section{Algorithm}
The OmniAssistant comprises three key components that build a bridge across numerous data modalities: Infinite Data Acquisition and Preprocessing, Data Fusion with mutual enhancement. It involves obtaining a variety of data, including video, audio, and text.
Fig.~\ref{fig:Algorithm} demonstrates the timeline-based data generation process conducted by OmniDataComposer's various components.

\begin{figure*}[htb]
\centering
\includegraphics[height=2.5in]{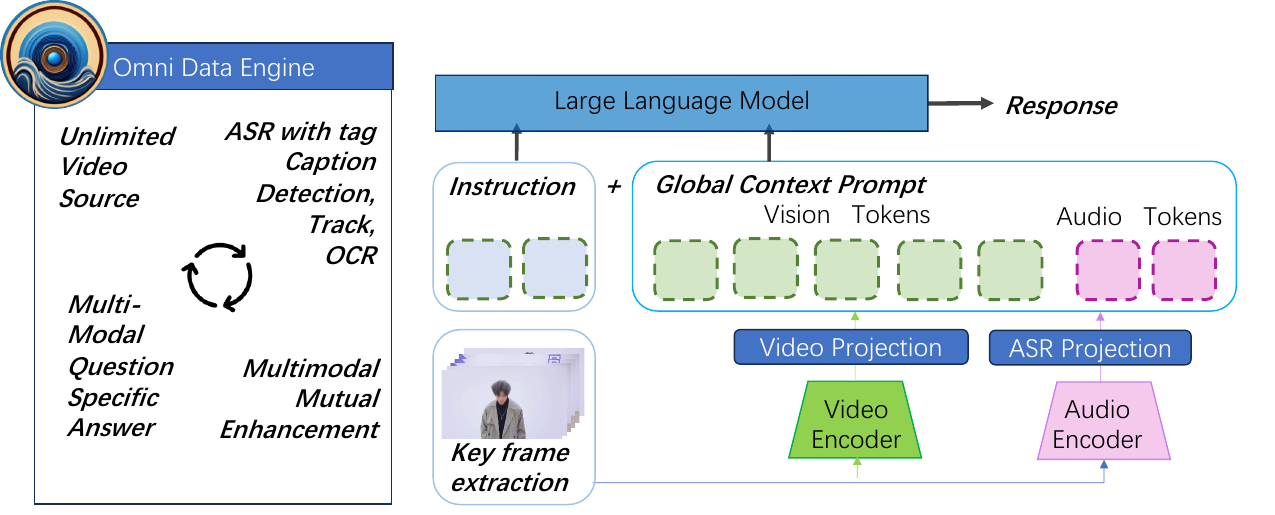}
\caption{Omni model architecture}
\label{figmodel}
\end{figure*}

\subsection{ASR and OCR} 
In video comprehension, ASR (Automatic Speech Recognition) and OCR (Optical Character Recognition) serve crucial functions, chiefly in the generation of textual information.
We use ASR to convert spoken language in the videos into written text, whereas OCR reads and digitizes any text present in the images or in the video frames.

ASR primarily transcribes the audio elements of videos, converting spoken language into text for additional processing and comprehension. By leveraging ASR, dialogues, speeches, or sound effects can be converted to a written format. 

Whisper-AT constitutes a combined audio tagging and speech recognition model. In addition to recognizing spoken text, it demonstrates the capability of identifying background noise within a solitary forward pass.

OCR's core function in video comprehension is the recognition and extraction of embedded textual information. Signs, subtitles, or screenshot captures can be subjected to OCR to retrieve textual content. This content can then undergo further analysis to discern the video's topic, categorise the video, or identify specific information such as names, locations, dates, etc.

The OCR system we employ, PaddleOCR, is capable of identifying and extracting textual information embedded within video frames. This system comprises two pivotal components: a model for text detection and another for text recognition. The recognition model of PP-OCRv3 utilises an input shape of [3, 48, 320]. Beyond the capabilities for Chinese and English, this system can be extended to perform text detection and recognition tasks for several other languages.

\subsection{Image/Dense Caption Extraction} 
To bridge the semantic gap between different modalities and enhance understanding of visual content, our algorithm strategically incorporates image and dense captioning techniques. Specifically, we make use of BLIP-2 \cite{li2023blip2} for image captioning and Shikra for dense captioning.

\subsubsection{Image Captioning with BLIP-2} Image captioning plays an essential role in elaborating visual data, and our choice of algorithm, BLIP-2 \cite{li2023blip2} , uniquely serves this purpose. BLIP-2 \cite{li2023blip2} , abbreviated from Bootstrapping Language-Image Pre-training with Frozen Image Encoders and Large Language Models, provides a new frontier of multimodal learning. This technique bridges the gap between image understanding and language generation, providing a contextual description for each image input.

With BLIP-2 \cite{li2023blip2} , each image input goes through an encoding process that generates a rich feature vector, capturing the essence of the image. This encoded information is then fed into a language model, which crafts a human-like, comprehensive description of the image. This neat interface of translating an image into a verbose description facilitates a clearer rendering of visual content into AI's language understanding pipeline.

\subsubsection{Dense Captioning with Shikra} While image captioning delivers a broad overview of an image, dense captioning drills down to the object level, providing granular insight into individual entities within the image. Our algorithm employs Shikra for this task.

Shikra \cite{chen2023shikra} is an innovative algorithm designed for generating dense captions. This model's novelty lies in its capability to identify multiple objects within an image and generate a caption for each object independently. Shikra \cite{chen2023shikra} achieves this by incorporating a detection network that spots the objects within an image, followed by a captioning network that generates a separate description for each identified object.

Bringing it all together, the combined use of BLIP-2 for image captioning and Shikra \cite{chen2023shikra} for dense captioning equips our OmniDataComposer with the competence to handle a wide span of complexities within video/image content, thereby enhancing cross-modal data understanding and manipulation.

\subsection{Object Detection and Object Tracking} 
Occupying a cornerstone position in our algorithm's Object Detection and Object Tracking section, AS-One functions as the main engine. Developed as a versatile, modular library in a Python wrapper, AS-One integrates various detection and tracking algorithms onto a unified platform with the core objective being the practical application of YOLO object detection and object tracking protocols.

This library plays a pivotal role in synchronizing diverse trackers, namely ByteTrack \cite{zhang2022bytetrack}, DeepSORT \cite{wojke2017simple} , and NorFair, and aligns them seamlessly with different YOLO versions. AS-One showcases broad compatibility achieved with minimal coding, underscoring its adaptability and flexibility.

AS-One exhibits remarkable proficiency in scenarios requiring object detection and tracking, particularly where the detection categories are predetermined, as evidenced in the 80 categories outlined in the COCO \cite{cocodata} dataset. The OmniDataComposer system deploys yolov7 \cite{wang2023yolov7} as its primary detector and ByteTrack \cite{zhang2022bytetrack} as its fundamental tracker, thereby ensuring seamless execution of expedited detection and comprehensive tracking throughout entirety of the video. The system also exhibits real-time tracking capabilities at the rate of 20fps.

\subsection{Recognize Anything Model(RAM)} 

The RAM \cite{zhang2023recognize} algorithm stands out with its capability of discerning and tagging an extensive range of entities within an image, presenting a comprehensive understanding of the intricacies involved in each visual input. The algorithm is rooted in exhaustive pre-training, assisting it in defining and labelling an astounding array of over 6400 object categories. This pre-training also empowers RAM \cite{zhang2023recognize} to confidently strut through the complexities of color, scene, and activity recognition, amplifying the scope of visual information processed.

RAM \cite{zhang2023recognize} has leveraged the advancements of deep learning and computer vision. This combination blesses RAM \cite{zhang2023recognize} with an exceptional ability to identify objects in images and mark them accordingly. This object-oriented approach provides a granular insight into the image content, offering a distinct perspective to machine learning algorithms.

Moreover, RAM exhibits a unique quality of providing automatic labels for the tagged entities. This automatic labelling makes the information more organized and easily accessible for further processing.

The integration of RAM widens the horizons to cross-modal data correction, allowing our model to identify possible mistakes in video transcription or image captioning and employ RAM's tagging capability to correct these errors.

In conclusion, prioritizing RAM within our OmniDataComposer has been instrumental in enhancing the robustness and reliability of our model. The algorithm brings forth unparalleled precision and awareness to multimodal learning, paving the way for a new era in the domain of artificial intelligence.

\begin{figure*}[ht]
\centering
\includegraphics[width=6.6in]{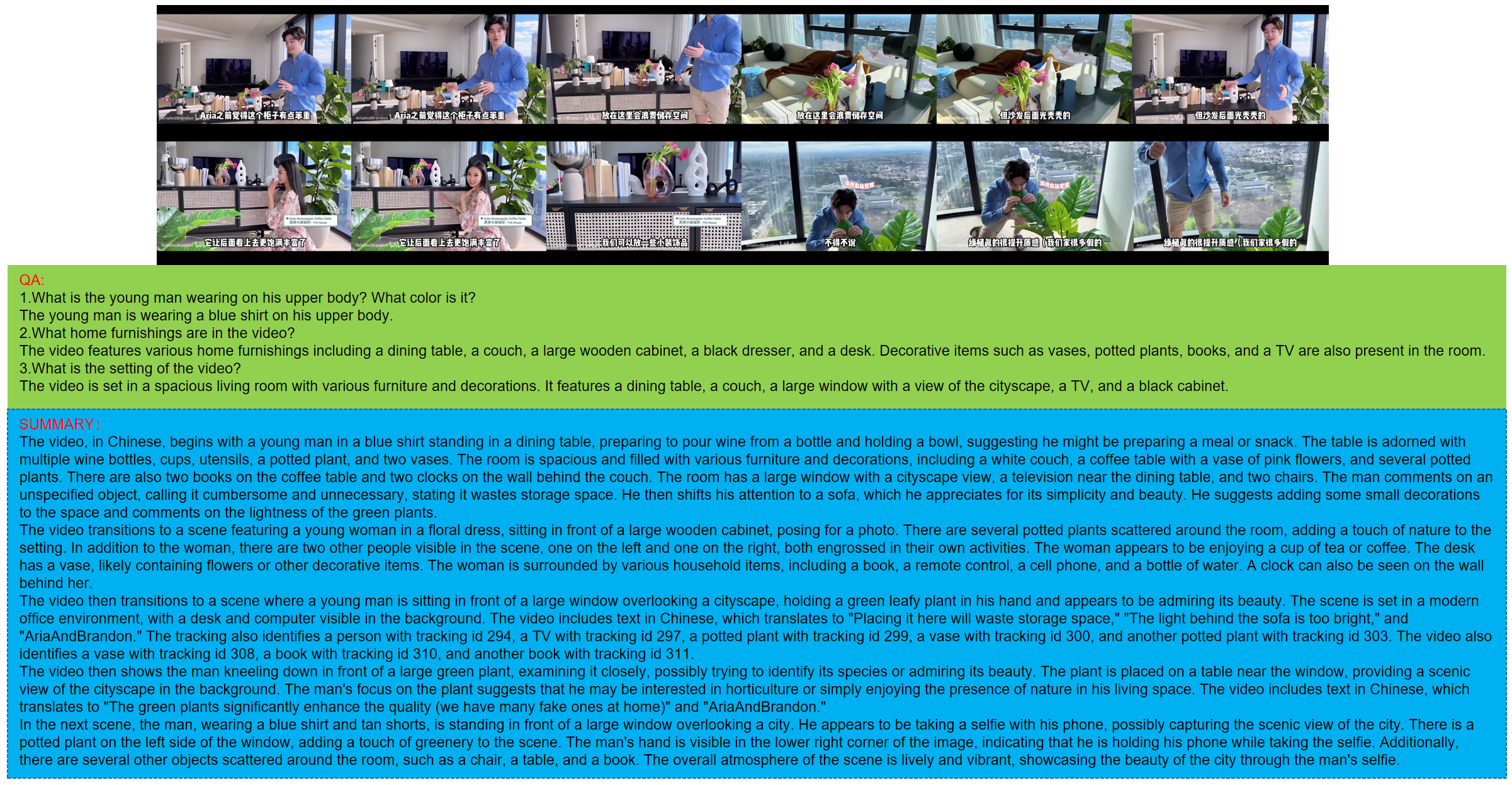}
\caption{Presentation of the Results Obtained from OmniLLM's Question-Answering and Summarization Tasks.}
\label{fig:summary}
\end{figure*}

\subsection{Multi-Modalities Mutual Enhancement}
Capitalizing on the synthesis of ChatGPT4 and Langchain, OmniDataComposer not only receives and consolidates multimodal datasets but also processes them efficiently. It navigates the transformation of temporally discrete multimodal data into exhaustive and detailed content of the video.

The notion of Multi-Modalities Mutual Enhancement (MMME) at the heart of the OmniDataComposer algorithm relies on the amalgamation and maneuvering of diverse data forms procured from several modalities, which include video, audio, and text. This synchronized process underpins the potential for mutual enhancement and improvement across the array of modalities.

Working on the hypothesis that the coordinated effort of various modalities eclipses the performance of independent modalities, MMME allows each data modality—video, audio, or text—to collate and interpret information gleaned from their counterparts, subsequently honing the perception and portrayal of data within each modality.

In the initial stages, the algorithm decodes the video content employing video/image caption extraction and object-tracking mechanisms. In tandem, Automatic Speech Recognition (ASR) converts the audio content from the video into a script, whereas Optical Character Recognition (OCR) extracts visual titles and textual content. This transformational stage refines raw video inputs into a structured configuration of multimodal data.

Following this, the independently extracted data from each modality intersects and mutually reinforces one another. The text end-product of ASR and OCR aids in the generation of more precise video captions, while concurrently, the derived video and image data amend any potential inaccuracies from the ASR and OCR transformations.

\subsection{Omni Model Architecture}
In Fig.~\ref{figmodel}, Omni model encodes each data modality into a format conducive to processing by the large language model. We use VideoMAE \cite{tong2022videomae} as video backbone. While for audio, we utilize transformer-based models akin to Whisper \cite{gong2023whisper}. The video encoder and audio encoder extract respective features into text-based embeddings.

The resulting encodings are then fused together to form a single, comprehensive representation of the data. This fusion process is handled by cross attention and projection that has been specially designed to handle diverse data types. This unified representation maintains the sequential nature of the data, ensuring the temporal dependencies of the original video are preserved.

Finally, the unified representation is fed into the large language model, which in our case is OmniLLM. OmniLLM processes instructions along with the vision and audio embeddings generating a sequential answer. OmniLLM can be used for various tasks, such as video captioning or question-answering based on video content. 

Through these steps, OmniAssistant successfully transforms multimodal data into a unified format, creating a coherent narrative from a video input and preparing it for high-level language model processing. It generates dataset and trains itself. This unique approach paves the way for a more integrated and contextually aware interpretation and generation of multimodal data.

\section{Experimental Results}
The innovative OmniDataComposer algorithm and the universal language model, OmniLLM, were evaluated based on their performance in answering video-based questions as well as generating in-depth, comprehensive video summaries.

OmniLLM accurately characterized and described the young man's attire, enumerated the various home furnishings visible in the video, and depicted the video's setting, evidencing its efficacy in analyzing and comprehending visual data.

Tasked with the objective of summarizing the video, OmniLLM provided a nuanced summary that recognized and articulated the sequences within the video: from the young man preparing a meal, transitioning to a scene featuring a young woman, up to a scene shift where a young man is seen admiring a leafy plant. The results of this summarization task are visually represented in Fig.~\ref{fig:summary}.

The summary incorporated information detailing the characters' actions, locations, sequence of events, and object identification. OmniLLM further demonstrated its capacity to detect, translate, and integrate Chinese text present in the video, thereby enhancing the granularity and quality of the summary.

OmniLLM demonstrates proficiency in object recognition and spatial-temporal tracking of these objects across video scenes whilst assigning correct tracking IDs. This evidence not only underscores OmniLLM’s capabilities in object recognition and tracking but also upholds the spatial and temporal integrity of the video data.

The experimental findings suggest that the innovative OmniDataComposer approach to multimodal data fusion significantly enhances OmniLLM’s capabilities to effectively comprehend image-caption and video-caption data. This foundation paves the way for producing higher-quality datasets for vQG, vQA, and refined question-answering tasks. Consequently, it signifies a substantial leap in multimodal learning, unlocking considerable potential for augmenting AI's comprehension and generation of intricate real-world data.

\begin{figure*}[ht]
\centering
\includegraphics[height=3.5in]{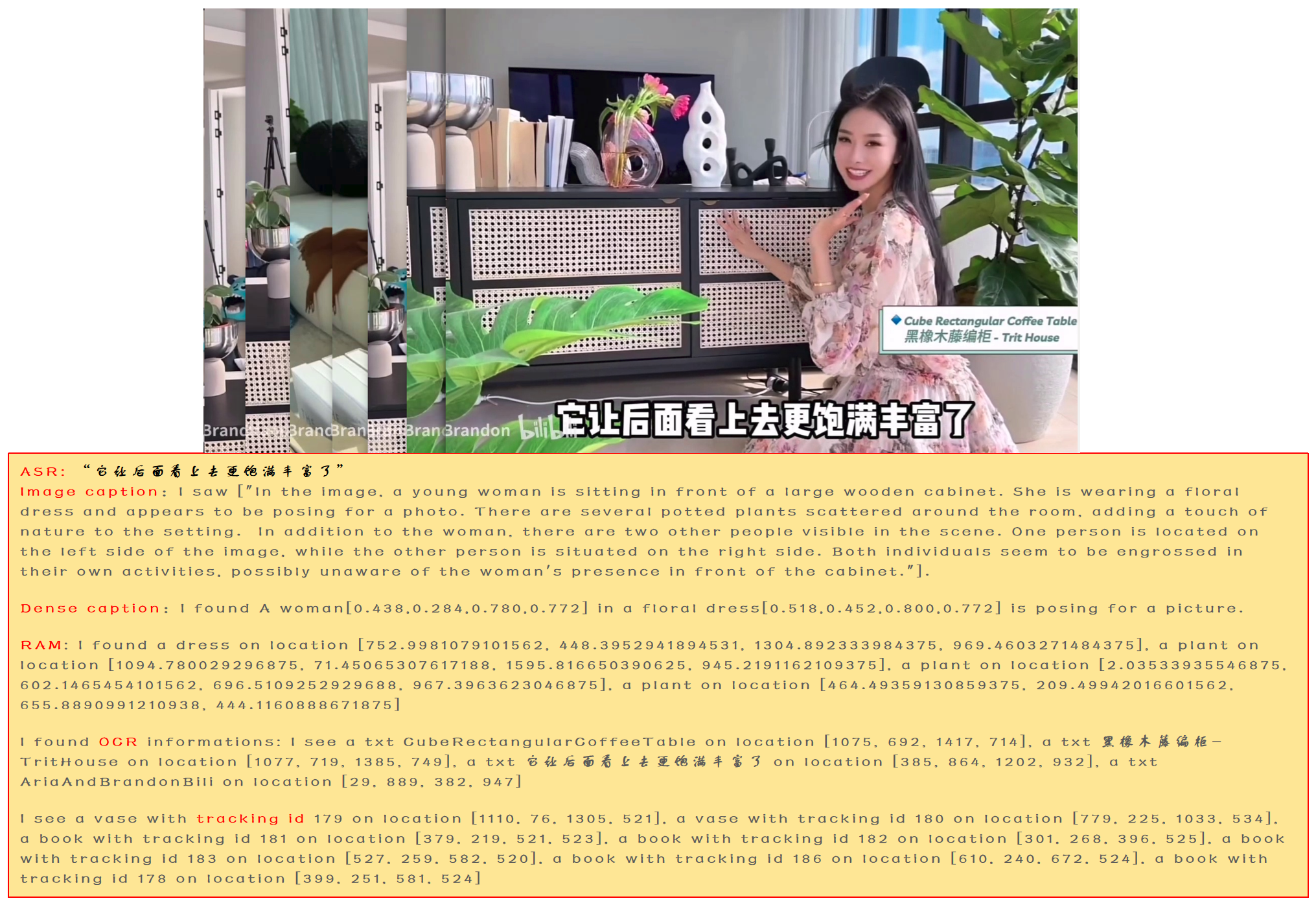}
\caption{Ablation Studies of OmniDataComposer Algorithm}
\label{fig:ablation}
\end{figure*}

\section{Ablation Studies}

The influence and contribution of various components (ASR, OCR, RAM, Video Captioning) of the OmniDataComposer model were analyzed by conducting a comprehensive set of ablation studies. We sequentially removed each component from the model to quantify the effect on the final output quality. Fig.~\ref{fig:ablation} presents the data pertaining to various components (ASR, OCR, RAM, Video Captioning) of the OmniDataComposer model. Each component has its own distinct data characteristics.

\subsection{Without ASR} The value of ASR, especially when implemented with Whisper-at, extends beyond merely transcribing human speech in videos. It is capable of recognizing and transcribing not only human speech but also background noises and animal sounds - aspects that are often overlooked but can offer pivotal contextual insights. In videos that lack embedded subtitles or where visuals do not provide complete context, audio cues can be crucial. Eliminating ASR results in the loss of this underlying layer of information, potentially comprising mood-evoking background scores, indicative sounds such as ambulance sirens, and animal sounds that may provide significant situational cues. This loss diminishes the depth of understanding and interpretation of the given scenario. The real-world applicability of ASR goes beyond interpreting speech; it provides an overarching conversion of sound to text, capturing a holistic acoustic representation in an analyzable data format. Thus, the absence of ASR can lead to a notable reduction in the richness of contextual understanding in the data representation, emphasizing its critical role in the seamless fusion of multimodal data.

\subsection{Without OCR} The role of Optical Character Recognition (OCR) in video analysis is pivotal as it extracts, interprets, and quantifies text embedded within video content, including various forms such as bullet screen comments, subtitles, captions, or text on objects such as billboards. OCR enables richer comprehension of videos by mining data that might be overlooked, such as texts displayed in the background, or brief subtitle cues, which often profoundly influence the context and interpretation of the video.

OCR acts as a bridge between image-based data (video frames) and language-based information (text within the video frames). When combined with Image and Dense Caption Extraction, OCR empowers language models to understand, interpret, and caption videos more accurately, thereby yielding a comprehensive understanding.

In instances where the video's audio component is absent, muted, or in an unfamiliar language, OCR becomes indispensable in deciphering text clues from the video, thus allowing viewers to follow and grasp the main arguments. Consequently, OCR proves vital in enhancing the quality of video analysis, ensuring precise analytics, and providing a comprehensive interpretation of the content.

\subsection{Without RAM} Our ablation studies furnished an avenue for assessing the functionality of OmniDataComposer in the absence of the Recognize Anything Model (RAM) \cite{zhang2023recognize} , a critical component in our algorithm. With the capacity to identify in excess of 6400 object categories, RAM \cite{zhang2023recognize} plays a pivotal role in earmarking and arranging various elements within each frame, substantially augmenting the efficacy of Dense Caption Extraction and Image Caption Extraction.

Due to the lack of RAM \cite{zhang2023recognize} , our system's proficiency in discerning and categorizing discrete objects within the video frame was significantly compromised. This deficiency constrained the level of detail in the narrative, as it lacked the object-specific data typically provided by RAM \cite{zhang2023recognize} , thereby cultivating a less sophisticated narrative.

\subsection{Without Video Captioning} Through our ablation studies, we assessed the efficacy of our proposed system, 'OmniDataComposer', without the integral components of video captioning, namely Dense Caption Extraction and Image Caption Extraction. Serving as the bedrock of our algorithm, Dense Caption Extraction offers a comprehensive, detailed narrative for every frame in a video, while Image Caption Extraction provides a general summarizing perspective of each scene.

In the absence of Dense Caption Extraction, our system experienced a substantial reduction in its ability to achieve in-depth comprehension of each frame in the video. The scarcity of microscopic focus resulted in a lesser detailed narrative, owing to the lack of dense, detail-rich captions typically provided by Dense Caption Extraction.

Elimination of Image Caption Extraction resulted in losing the summarizing vision of each scene. This led to a significant shortfall, resulting in a compromised holistic understanding of each frame, given that this method constructs a wider picture which mitigates the risk of losing oneself in excessive details.

The joint absence of these two techniques significantly impaired the output quality of OmniDataComposer, thus emphasizing their indispensable roles in the superior performance of our system in terms of video content comprehension.

\section{Conclusion and Future Works}
We demonstrated the potency of the OmniDataComposer in this paper, an innovative method that transforms the orchestration of multimodal data, thereby refining and optimizing the interaction between video, audio, and text modalities. Our algorithm effectively mastered operations like video/image caption extraction, dense caption extraction, ASR, OCR, title association, and object tracking. Our methodology was capable of identifying over 6400 categories of objects and bridging different modalities for a rounded enhancement, creating extensive narratives from each video input, thereby facilitating their processing by large language models.

Moving forward, our future research will mainly focus on optimizing the datasets for each modality to further enhance the capability of the OmniDataComposer in unlimited data generation. Our vision is to build a robust base that can provide invaluable insights for models like ChatGPT, with a primary goal of improving the quality of datasets used for video captioning and simplifying question-answering tasks based on video content. OmniDataComposer marks the beginning of a new era in multimodal learning, and we will continue refining its functionalities and algorithms to assimilate and generate a larger scale and more complex real-world data, thus paving the way for significant advancements in AI's understanding and data generation capabilities.

{\small
\bibliographystyle{ieee_fullname}
\bibliography{egbib}
}

\end{document}